# Accuracy-Preserving Stability Regularization for Large-Scale Retail Demand Forecasting

Accuracy-Preserving Stability Regularization

Jize Li[1*], Jiani He[2†], Dishu Yang[3†], Dingyan Shang[4], Jingjing Liu[5], and Shiqi Huang[6]

[1] Boston University, Boston, MA, USA; [2] Massachusetts Institute of Technology, Cambridge, MA, USA; [3] Northeastern University, San Jose, CA, USA

[4] University of Southern California, Los Angeles, CA, USA; [5] University of California, Berkeley, Berkeley, CA, USA; [6] Washington University in St. Louis, St. Louis, MO, USA

jizel@bu.edu; jianihe@alum.mit.edu; yang.dis@northeastern.edu; dingyanshang@gmail.com; mayliujj@berkeley.edu; juliahuangsq01@gmail.com

* Corresponding author: Jize Li, jizel@bu.edu. † Jiani He and Dishu Yang contributed equally and share second authorship.

**Abstract**

Retail demand forecasts are reused across replenishment, capacity, labor, and transportation planning cycles. Point-error objectives do not constrain abrupt movement between adjacent forecasts, while post-hoc smoothing acts only after model fitting. We ask whether a training-time penalty on consecutive within-series movement can improve horizontal forecast-path stability without materially changing point accuracy. The penalty is evaluated in a temporal-structured pipeline combining recent-demand embeddings with calendar, price, hierarchy, item, and store features. On selected M5 demand series at 1000, 3000, and 4000-series scales, the stability-aware hybrid model improves Forecast Stability Score over XGBoost by 6.91%, 6.66%, and 7.68%, respectively, while RMSE changes remain within 0.72% across three random seeds. Post-hoc exponential smoothing attains lower raw movement but incurs a larger RMSE cost; training-time regularization preserves more point accuracy and performs favorably under normalized stability. These findings extend forecast evaluation from point-error minimization toward an accuracy-stability trade-off perspective for operational retail forecasting.

## 1 INTRODUCTION

Demand forecasts are working inputs to replenishment, inventory, production, labor, and transportation decisions [1, 2, 3]. In repeated planning cycles, abrupt adjacent forecast movement can trigger order revisions, capacity changes, rescheduling, and upstream adjustments even when average error remains acceptable. The practical issue is therefore not accuracy alone, but whether forecasts remain usable as plans are revisited. This concern is related to forecast stability and demand amplification [4, 5, 6], and is especially visible in intermittent retail series with zeros, sparse spikes, promotions, and price changes [7, 8, 9].

Most forecasting systems are selected with point-error criteria, which reward closeness to realized demand but do not restrict movement along the forecast path [3, 6]. Temporal-only models may omit structured retail signals; tabular models based on lags and rolling statistics may not learn compact recent-demand representations. Post-hoc smoothing reduces movement only after fitting and may suppress responsive variation. M5 makes these trade-offs testable by pairing item-store histories with calendar and sell-price tables [9]. We therefore use a lightweight multiscale temporal encoder with XGBoost for heterogeneous operational features [15].

This study asks whether horizontal forecast-path stability can be incorporated into model training without materially degrading point accuracy, and whether such training-time regularization offers a different accuracy-stability operating point than post-hoc smoothing. Stability regularization, rather than the model combination itself, is the central issue. The temporal-structured pipeline is used to test that question under realistic retail covariates.

The paper makes three contributions. First, it adapts a training-time, same-series total-variation penalty for consecutive forecasts to an XGBoost objective and contrasts it with post-hoc smoothing. Second, it uses a temporal-structured retail pipeline that combines a compact demand-history embedding with calendar, price, hierarchy, item, and store features. Third, it evaluates point accuracy, raw and normalized stability, smoothing baselines, seed robustness, leakage, and runtime on consistent M5 subsets. The theoretical contribution is to formalize horizontal forecast-path stability as a decision-relevant evaluation dimension alongside point accuracy, and to position training-time regularization and post-hoc smoothing as distinct operating points in an accuracy-stability trade-off. This is not a state-of-the-art accuracy claim.

## 2 RELATED WORK

### 2.1 Deep Time Series Forecasting

Recent deep time-series models improve temporal representation and point accuracy through multiscale mixing, patching, efficient attention, decomposition, and MLP-based mixing [10, 11, 12, 13, 14]. They generally do not treat adjacent forecast-path movement as an explicit operational objective.

Retail forecasting also depends on calendar, price, event, product, and store signals [2, 9]. Standard accuracy-oriented training does not encode a preference for restrained adjacent movement [3, 6]. Here, temporal representation learning supports the analysis, while operational stability remains the main object of study.

### 2.2 Supply Chain and Retail Demand Forecasting

Retail demand forecasting involves sparse SKU-store series, seasonality, prices, promotions, and hierarchy. Forecasting competitions shaped empirical practice [16], and M5 provides large-scale Walmart sales, calendar, and sell-price data [9]. Practical retail methods must use operational covariates and be evaluated in decision-relevant contexts [1, 2, 3].

Tree-based learning remains practical for nonlinear, heterogeneous demand drivers [2, 9]. XGBoost provides scalable boosted trees for sparse features and interactions [15], motivating hybrids that add learned demand-history representations to tabular covariates.

### 2.3 Hybrid Forecasting Frameworks

The hybrid pipeline combines a small multiscale MLP inspired by TimeMixer [10] with XGBoost over price, calendar, event, hierarchy, item, and store variables [15]. The encoder is lightweight and is not a full TimeMixer reproduction.

This pipeline is the test bed, not the claimed novelty: the same structured learner is compared with and without stability regularization while retaining M5 covariates [9].

### 2.4 Forecast Stability and Robust Forecasting

Forecast stability has several meanings. Godahewa et al. distinguish horizontal stability across adjacent target dates from vertical stability across successive forecast origins for the same target [6]. Van Belle et al. optimize rolling-origin stability through a composite neural loss [22], and Caljon et al. study dynamic weighting of that objective [23]. Supply-chain work links forecast revisions and method choice to demand amplification [4, 5, 17], while intermittent-demand research connects forecasts to inventory performance [7, 8, 18].

The contribution is therefore not the invention of stability-aware loss optimization. It studies horizontal path regularization in a boosted-tree retail pipeline, with post-hoc smoothing and scale-normalized stability as explicit comparisons.

## 3 METHODOLOGY

### 3.1 Problem Formulation

Let $y_{i,t}$ be demand for item-store series i at time t. Given historical demand $y_{i,t-L:t-1}$ and structured covariates $x_{i,t}$ containing calendar, price, event, category, and lag-derived information, the task is to estimate $\hat{y}_{i,t}$.

### 3.2 Temporal Representation Learning and Feature Fusion

The temporal encoder maps historical demand sequences to compact embeddings:

$$z_{i,t} = g_\theta(y_{i,t-L:t-1}) \quad (1)$$

where $g_\theta(\cdot)$ is a lightweight multiscale MLP feature extractor inspired by TimeMixer [10], not a full reproduction. It transforms a shifted 28-day history using raw lags, first differences, and pooled means and standard deviations at 2-, 4-, 7-, and 14-day resolutions. The final hidden layer yields a 16-dimensional embedding $z_{i,t}$.

The forecasting model concatenates temporal embeddings with lag, rolling, calendar, price, hierarchy, item, and store features:

$$h_{i,t} = [z_{i,t}; x_{i,t}] \quad (2)$$

and passes the fused representation to a gradient-boosted forecasting layer [15]:

$$\hat{y}_{i,t} = f_\varphi(h_{i,t}) \quad (3)$$

The encoder's scaler and MLP are estimated only from training-window sequences, using a seeded training-only sample capped at 50,000 windows for lightweight learning. The fitted encoder generates embeddings for all rows; validation uses only train-fitted transformations. Lag and rolling variables use shifted demand, and the boosted-tree experiments retain all selected training rows.

### 3.3 Stability-Aware Objective

The forecast loss is mean squared error:

$$L_{forecast} = (1/N) \sum_{i,t}(y_{i,t} - \hat{y}_{i,t})^2 \quad (4)$$

The stability loss penalizes consecutive forecast changes within each demand series:

$$L_{stability} = (1/N_s) \sum_i \sum_{t>1} |\hat{y}_{i,t} - \hat{y}_{i,t-1}| \quad (5)$$

where $N_s$ is the number of valid adjacent same-series prediction pairs. Equation (5) is a series-separable discrete total-variation penalty: it couples neighboring dates within one item-store series, not unrelated series. The total objective is

$$L_{total} = L_{forecast} + \lambda L_{stability} \quad (6)$$

where $\lambda$ controls the stability-accuracy trade-off. We evaluate $\lambda \in \{0, 0.01, 0.05, 0.1, 0.5\}$ and use $\lambda = 0.05$ for the final model.

### 3.4 Implementation of the Stability-Aware XGBoost Objective

The XGBoost custom objective uses row-level gradients and Hessians. The MSE term contributes gradient $2(\hat{y}_{i,t} - y_{i,t})$ and Hessian 2. Stability pairs are formed after sorting training rows by item-store series and day index, using only consecutive rows from the same series.

Each adjacent pair contributes a sign subgradient to the later prediction and the opposite contribution to the preceding prediction; boundary rows receive one contribution. The sign term uses $\max(|\Delta|, \varepsilon)$, and a small $\lambda\varepsilon$ term keeps the Hessian positive. Validation and test targets are excluded. The objective therefore encodes restrained same-series movement inside empirical risk rather than through global or post-hoc smoothing. The leakage audit verified chronological splitting, shifted features, train-fitted encoders, and training-only pairs.

### 3.5 Post-Hoc Smoothing Baseline

To separate training-time stability regularization from post-hoc smoothing, we also evaluate exponential smoothing applied after prediction, following the broad use of exponential-smoothing ideas in forecasting and robust forecasting [19, 3]:

$$\tilde{y}_{i,t} = \alpha\, \hat{y}_{i,t} + (1 - \alpha)\, \tilde{y}_{i,t-1} \quad (7)$$

with $\alpha \in \{0.5, 0.7, 0.9\}$. Smoothing is applied independently within each demand series and only to model predictions. It does not change the fitted forecasting model or use validation targets.

### 3.6 Evaluation Metrics

We report MAE, RMSE, WAPE, FSS, normalized FSS (nFSS), volatility, runtime, and relative changes against XGBoost. Following forecasting-metric guidance [20, 21, 3], WAPE is

$$\mathrm{WAPE} = 100 \times \left(\sum_{i,t} |y_{i,t} - \hat{y}_{i,t}|\right) / \left(\sum_{i,t} |y_{i,t}|\right) \quad (8)$$

FSS is defined as

$$\mathrm{FSS} = (1/N_s) \sum_i \sum_{t>1} |\hat{y}_{i,t} - \hat{y}_{i,t-1}| \quad (9)$$

Lower FSS indicates smoother consecutive forecasts. In the terminology of forecast stability [6], it is horizontal: adjacent target-date predictions are compared within one horizon. Vertical instability instead compares revisions for the same target across forecast origins [22, 23]. We also report a scale-normalized score:

$$\mathrm{nFSS}_i = \left[(1/(T_i-1)) \sum_{t>1} |\hat{y}_{i,t} - \hat{y}_{i,t-1}|\right] / \left[(1/T_i) \sum_t |y_{i,t}| + \varepsilon\right] \quad (10)$$

$$\mathrm{nFSS} = (1/M) \sum_{i=1}^{M} \mathrm{nFSS}_i \quad (11)$$

where $\varepsilon = 10^{-6}$. Low-volume series can still yield large nFSS values. The results therefore concern within-horizon path smoothness, not rolling-origin revision stability.

## 4 EXPERIMENTS

### 4.1 Dataset and Setup

We use the M5 dataset, which provides daily Walmart item-store demand with calendar and sell-price information [9]. The main study uses 1000, 3000, and 4000 series under the same boosted-tree configuration: all selected training rows, 300 XGBoost trees, identical features, fixed temporal-encoder settings, and $\lambda = 0.05$.

The split is chronological, with all training dates before validation dates. Features include shifted lag and rolling demand, calendar, sell-price, and categorical item/store variables. Repeated experiments use seeds 42, 123, and 2024. An automated leakage audit also checked split integrity, train-only preprocessing, and same-series training-only stability pairs.

### 4.2 Metric Correction for Intermittent Demand

We remove raw MAPE because M5 has many zero and near-zero demand observations [9]. MAPE becomes undefined at zero and unstable near zero, a known issue for intermittent demand [21, 8, 3]. We report WAPE instead. Because low-volume series can make nFSS large in absolute value, we interpret nFSS as a relative robustness metric, not an operational threshold.

### 4.3 Baselines

We compare XGBoost, encoder-only forecasting, the encoder + XGBoost hybrid, and the stability-aware hybrid model. In the tables and figures, “TimeMixer” or “TimeMixer only” denotes the lightweight encoder described in Section 3.2, not a full TimeMixer reproduction; “Hybrid + Stable” denotes the stability-aware hybrid model. The post-hoc smoothing baselines (“XGBoost + Smooth” and “Hybrid + Smooth”) use $\alpha \in \{0.5, 0.7, 0.9\}$.

### 4.4 Main Results

Table 1 reports mean ± standard deviation across seeds, and Figure 1 summarizes RMSE and FSS. XGBoost remains the strongest point-accuracy baseline, while the stability-aware hybrid model gives the lowest FSS among primary training methods. The result is an accuracy-stability trade-off rather than pure RMSE dominance.

Table 1. Mean and standard deviation across three random seeds under consistent full training configurations. WAPE is reported as the percentage error metric for intermittent retail demand.

| Series | Method | Runs | MAE | RMSE | WAPE (%) | FSS | Volatility |
|---|---|---|---|---|---|---|---|
| 1000 | XGBoost | 3 | 0.926 ± 0.034 | 1.774 ± 0.185 | 70.02 ± 4.82 | 0.220 ± 0.028 | 0.313 ± 0.043 |
| 1000 | TimeMixer | 3 | 0.970 ± 0.044 | 1.930 ± 0.316 | 73.28 ± 4.30 | 0.347 ± 0.076 | 0.391 ± 0.073 |
| 1000 | Hybrid | 3 | 0.926 ± 0.034 | 1.781 ± 0.198 | 70.02 ± 4.76 | 0.222 ± 0.030 | 0.315 ± 0.044 |
| 1000 | Hybrid + Stable | 3 | 0.927 ± 0.035 | 1.787 ± 0.208 | 70.08 ± 4.71 | 0.205 ± 0.030 | 0.302 ± 0.042 |
| 3000 | XGBoost | 3 | 0.953 ± 0.006 | 1.852 ± 0.081 | 69.81 ± 1.47 | 0.228 ± 0.007 | 0.326 ± 0.012 |
| 3000 | TimeMixer | 3 | 0.993 ± 0.012 | 2.002 ± 0.123 | 72.79 ± 1.37 | 0.401 ± 0.040 | 0.439 ± 0.027 |
| 3000 | Hybrid | 3 | 0.953 ± 0.007 | 1.850 ± 0.078 | 69.84 ± 1.43 | 0.229 ± 0.008 | 0.328 ± 0.013 |
| 3000 | Hybrid + Stable | 3 | 0.954 ± 0.007 | 1.854 ± 0.082 | 69.91 ± 1.45 | 0.213 ± 0.009 | 0.315 ± 0.012 |
| 4000 | XGBoost | 3 | 0.955 ± 0.013 | 1.876 ± 0.012 | 69.53 ± 1.00 | 0.232 ± 0.007 | 0.331 ± 0.006 |
| 4000 | TimeMixer | 3 | 0.998 ± 0.020 | 2.014 ± 0.037 | 72.65 ± 1.34 | 0.350 ± 0.027 | 0.399 ± 0.031 |
| 4000 | Hybrid | 3 | 0.956 ± 0.012 | 1.882 ± 0.012 | 69.56 ± 0.94 | 0.231 ± 0.005 | 0.330 ± 0.005 |
| 4000 | Hybrid + Stable | 3 | 0.956 ± 0.012 | 1.881 ± 0.014 | 69.58 ± 0.96 | 0.214 ± 0.006 | 0.318 ± 0.007 |

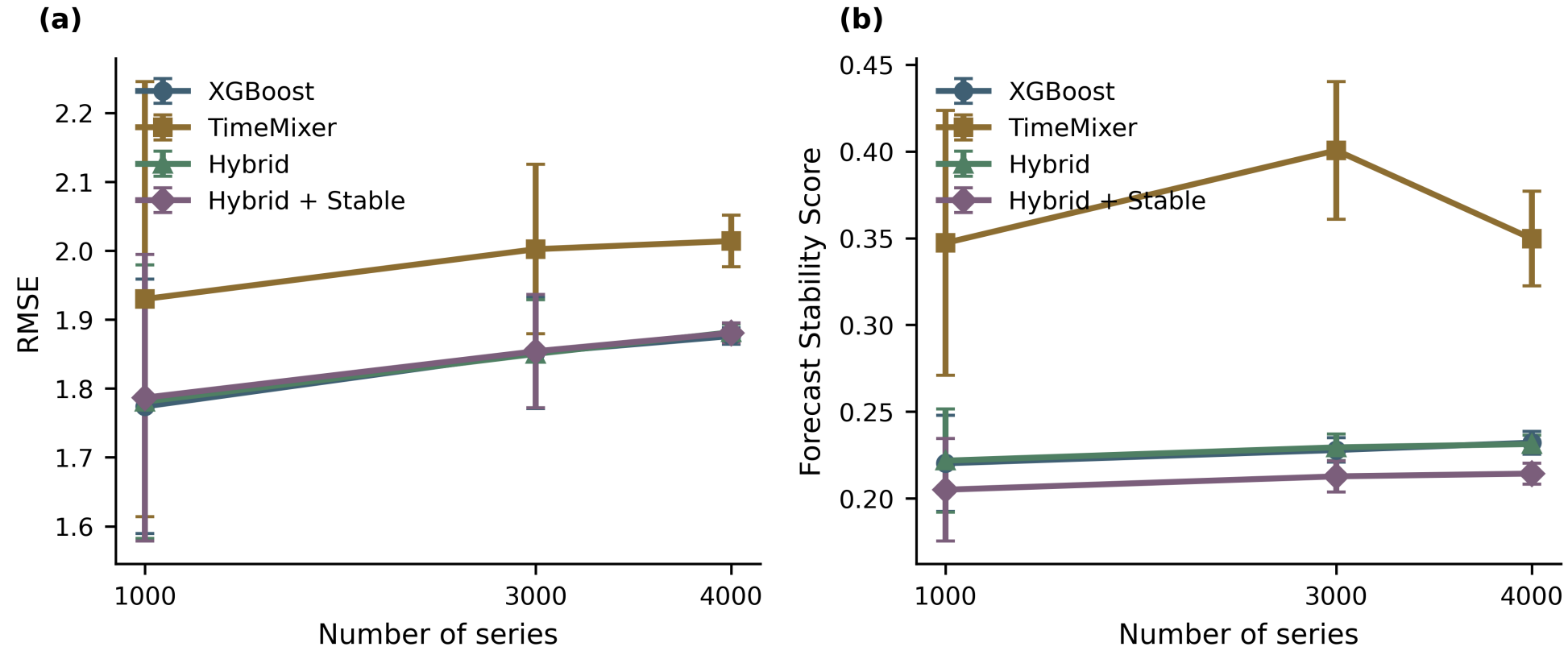

Figure 1. Main results under consistent full training configurations. Error bars denote standard deviation across three random seeds.

### 4.5 Stability-Accuracy Trade-off

The central empirical pattern is consistent across scales. As shown in Table 2, the stability-aware hybrid model improves FSS over XGBoost by 6.91%, 6.66%, and 7.68% at 1000, 3000, and 4000 series, respectively. Across these scales, the largest absolute RMSE change is 0.72%. Figure 2 places the methods on the accuracy-stability plane and shows that training-time regularization moves the operating point toward smoother forecasts while keeping RMSE close to the structured baseline.

Table 2. Relative stability, accuracy, and volatility changes for the stability-aware hybrid model versus XGBoost.

| Series | FSS Improvement (%) | RMSE Change (%) | Volatility Improvement (%) | Seeds Improved | Runs |
|---|---|---|---|---|---|
| 1000 | 6.91 | 0.72 | 3.77 | 3 | 3 |
| 3000 | 6.66 | 0.10 | 3.18 | 3 | 3 |

| Series | FSS Improvement (%) | RMSE Change (%) | Volatility Improvement (%) | Seeds Improved | Runs |
|---|---|---|---|---|---|
| 4000 | 7.68 | 0.24 | 3.85 | 3 | 3 |

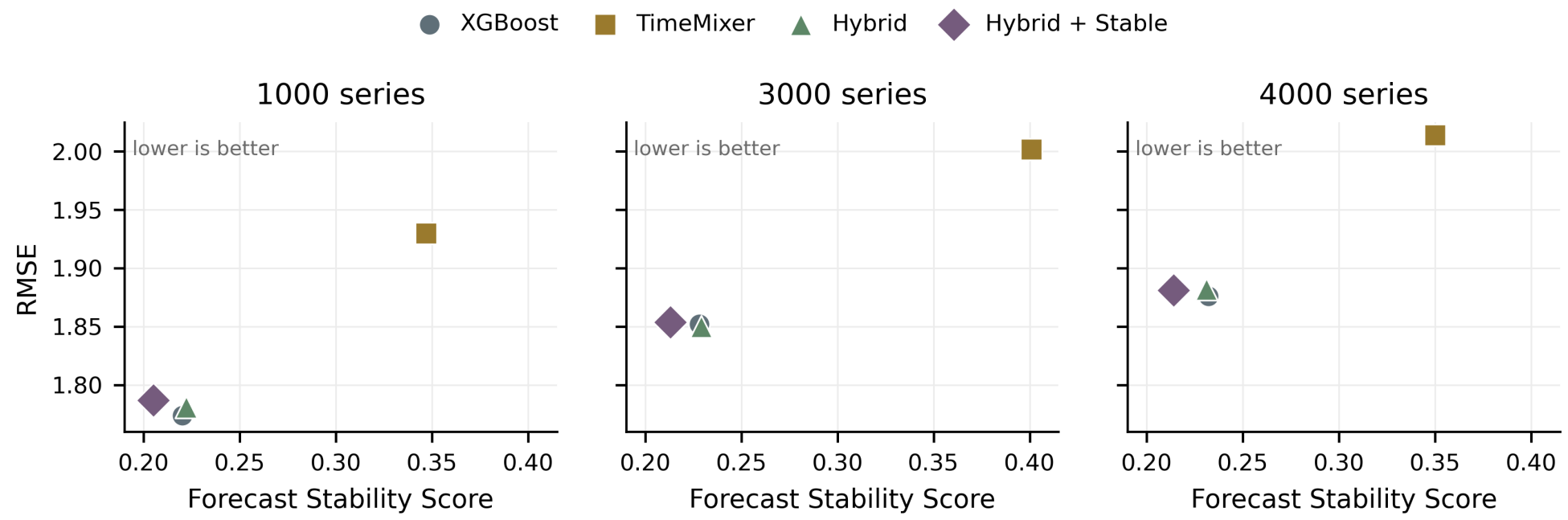

Figure 2. Accuracy-stability trade-off under the revised consistent scalability setup. Points show method-level operating positions at each scale.

### 4.6 Smoothing Baselines and Normalized Stability

Table 3 compares training-time regularization with the strongest post-hoc smoothing setting from the tested grid. The comparison reveals a clear trade-off. With α = 0.5, exponential smoothing gives the lowest raw FSS because it directly damps consecutive prediction changes, but at a cost in RMSE. Hybrid + Smooth increases RMSE by roughly 1.66–2.33% relative to XGBoost, while Hybrid + Stable remains within the reported bound of 0.72%.

The same pattern appears in the accuracy-stability trade-off: post-hoc smoothing reduces raw movement more aggressively, while the stability-aware hybrid model preserves more accuracy-oriented forecast structure.

Table 3. Post-hoc smoothing baseline comparison. Smoothing rows use the best raw-FSS setting from α ∈ {0.5, 0.7, 0.9}; lower FSS and nFSS are better.

| Series | Method | α | RMSE | WAPE (%) | FSS | nFSS | RMSE Δ (%) |
|---|---|---|---|---|---|---|---|
| 1000 | Hybrid + Stable | – | 1.787 ± 0.208 | 70.08 ± 4.71 | 0.205 ± 0.030 | 388.7 ± 113.8 | 0.72 |
| 1000 | XGBoost + Smooth | 0.5 | 1.812 ± 0.217 | 70.54 ± 4.56 | 0.116 ± 0.015 | 449.0 ± 132.9 | 2.15 |
| 1000 | Hybrid + Smooth | 0.5 | 1.815 ± 0.223 | 70.53 ± 4.55 | 0.116 ± 0.015 | 423.4 ± 94.7 | 2.33 |
| 3000 | Hybrid + Stable | – | 1.854 ± 0.082 | 69.91 ± 1.46 | 0.213 ± 0.008 | 373.5 ± 56.1 | 0.10 |
| 3000 | XGBoost + Smooth | 0.5 | 1.887 ± 0.090 | 70.41 ± 1.43 | 0.118 ± 0.004 | 452.1 ± 90.3 | 1.87 |
| 3000 | Hybrid + Smooth | 0.5 | 1.883 ± 0.087 | 70.41 ± 1.39 | 0.119 ± 0.004 | 437.3 ± 50.0 | 1.66 |
| 4000 | Hybrid + Stable | – | 1.881 ± 0.014 | 69.58 ± 0.96 | 0.214 ± 0.006 | 377.5 ± 73.7 | 0.24 |
| 4000 | XGBoost + Smooth | 0.5 | 1.911 ± 0.016 | 70.12 ± 0.92 | 0.121 ± 0.004 | 416.9 ± 74.6 | 1.88 |
| 4000 | Hybrid + Smooth | 0.5 | 1.917 ± 0.021 | 70.15 ± 0.88 | 0.120 ± 0.003 | 423.3 ± 90.1 | 2.16 |

Scale-normalized stability supports the same interpretation: the stability-aware hybrid model improves nFSS over XGBoost by 40.21%, 45.66%, and 41.93% at 1000, 3000, and 4000 series and outperforms the best Hybrid + Smooth baseline on nFSS. Because nFSS divides by average realized demand, we interpret it as a relative metric rather than an operational threshold.

### 4.7 Scalability and Runtime Analysis

Table 4 reports runtime under consistent full-row, 300-tree configurations. Relative to XGBoost, the stability-aware hybrid model incurs multipliers of 4.39×, 7.71×, and 9.72× at 1000, 3000, and 4000 series. Post-hoc smoothing adds little processing time after prediction but does not reduce the cost of fitting the underlying models.

Table 4. Runtime scalability under the revised consistent configuration. All main scales use full selected training rows and 300 XGBoost trees.

| Series | Method | Runtime Mean (s) | Runtime Std (s) | Peak Memory (MB) |
|---|---|---|---|---|
| 1000 | XGBoost | 17.31 | 6.53 | 3147.36 |
| 1000 | TimeMixer | 27.24 | 4.28 | 3147.36 |
| 1000 | Hybrid | 47.91 | 9.83 | 3147.36 |
| 1000 | Hybrid + Stable | 75.90 | 13.97 | 3147.36 |
| 3000 | XGBoost | 45.77 | 4.61 | 2893.42 |
| 3000 | TimeMixer | 187.26 | 21.47 | 2893.42 |
| 3000 | Hybrid | 269.29 | 38.70 | 2893.42 |
| 3000 | Hybrid + Stable | 352.97 | 39.81 | 2893.42 |
| 4000 | XGBoost | 53.09 | 10.63 | 3352.53 |
| 4000 | TimeMixer | 263.15 | 41.01 | 3352.53 |
| 4000 | Hybrid | 372.59 | 73.37 | 3352.53 |
| 4000 | Hybrid + Stable | 515.96 | 61.28 | 3352.53 |

### 4.8 Robustness Analysis

Across seeds, the stability-aware hybrid model improves FSS over XGBoost at all revised main scales. Averaged over 1000, 3000, and 4000 series, FSS improves by 7.08%, while the largest absolute RMSE change remains 0.72%. Additional diagnostics were conducted for internal validation but are omitted from the submission manuscript due to page limits.

## 5 DISCUSSION

Objective-based regularization does not dominate smoothing in every setting. When raw forecast movement is the sole concern, exponential smoothing is a strong and inexpensive baseline once predictions have been generated. In these experiments it produces the lowest raw FSS. That result defines a useful alternative operating point rather than weakening the comparison.

Training-time regularization occupies a different part of the trade-off. It gives up some raw smoothness relative to post-hoc smoothing, but it retains more of the accuracy-oriented forecast structure and performs better under scale-normalized stability. This distinction is operationally relevant because forecast changes can propagate into replenishment and upstream planning variability [4, 5, 17]. A planner may reasonably prefer moderate stabilization that remains responsive to demand over stronger damping that moves farther from the point-accuracy baseline.

### 5.1 Theoretical Implications

Point accuracy alone is incomplete when forecasts are reused in repeated decisions: two models with similar RMSE can impose different levels of forecast-path movement on planners. Horizontal stability is therefore a distinct evaluation dimension from conventional point error and from vertical rolling-origin revisions [6, 22, 23]. Training-time regularization and post-hoc smoothing should likewise be understood as different operating points. Smoothing can reduce raw movement more aggressively, whereas regularization can preserve more accuracy-oriented structure. This framing supports decision-aware model selection, but it does not establish inventory savings because the experiments evaluate forecast-level rather than downstream outcomes.

The runtime cost is real: temporal embedding extraction and stability-aware boosted-tree training make the final method slower than XGBoost, with multipliers of 4.39×, 7.71×, and 9.72×. These values reflect end-to-end experimental cost, not deployment-time inference latency.

Many supply-chain forecasts are generated in scheduled batch planning cycles rather than latency-critical online inference pipelines. Additional training-time cost may therefore be acceptable when reduced forecast churn lowers downstream planning instability. Still, this study profiles end-to-end experimental runtime rather than separating training-time and inference-time costs. Future work should report deployment-time inference profiles, including temporal embedding extraction and XGBoost prediction costs under realistic production batch sizes.

## 6 LIMITATIONS

The evaluation is limited to selected M5 subsets rather than the full hierarchy. Although M5 captures sparse item-store demand, prices, events, and hierarchy, it cannot represent every supply-chain environment. The largest full-configuration main experiment contains 4000 series; a capped 5000-series diagnostic was used only for internal validation and not for the main scalability claims.

FSS and nFSS measure horizontal movement across adjacent target dates, not vertical forecast revisions across rolling origins. Low-volume intermittent series can also yield large nFSS values, so the score is interpreted comparatively rather than as an operational threshold.

Finally, the evidence remains forecast-level rather than inventory-level. No replenishment simulation establishes changes in service levels, stockouts, holding costs, or schedule revisions. These outcomes, together with rolling-origin evaluation and broader retail datasets, are priorities for future work.

## 7 CONCLUSION

This paper examined whether horizontal forecast-path stability can be regularized during training without materially changing point accuracy. Across 1000, 3000, and 4000 M5 series, the same-series movement penalty improves FSS while keeping RMSE changes small. Post-hoc smoothing remains preferable when raw smoothness alone is the priority; training-time regularization offers a more accuracy-preserving operating point under scale-normalized stability. The evidence supports including stability alongside accuracy in retail forecast evaluation, while the limits of one dataset, one held-out horizon, and forecast-level rather than inventory-level outcomes remain important.